\documentclass[sigconf]{aamas} 

\usepackage{balance} 
\usepackage{algorithm}
\usepackage{algorithmic}
\usepackage{mathrsfs}
\usepackage{caption}
\usepackage{pifont}
\usepackage{subfigure}
\usepackage{fancyhdr} 
\usepackage{threeparttable}
\usepackage{tabularx}
\setcopyright{ifaamas}
\acmConference[AAMAS '23]{Proc.\@ of the 22nd International Conference
on Autonomous Agents and Multiagent Systems (AAMAS 2023)}{May 29 -- June 2, 2023}
{London, United Kingdom}{A.~Ricci, W.~Yeoh, N.~Agmon, B.~An (eds.)}
\copyrightyear{2023}
\acmYear{2023}
\acmDOI{}
\acmPrice{}
\acmISBN{}

\acmSubmissionID{179}

\title[AAMAS-2023 Formatting Instructions]{Accelerating Self-Imitation Learning from Demonstrations via Policy Constraints and Q-Ensemble}

\begin{abstract}
Deep reinforcement learning (DRL) provides a new way to generate robot control policy. However, the process of training control policy requires lengthy exploration, resulting in a low sample efficiency of reinforcement learning (RL) in real-world tasks. Both imitation learning (IL) and learning from demonstrations (LfD) improve the training process by using expert demonstrations, but imperfect expert demonstrations can mislead policy improvement. Offline to Online reinforcement learning requires a lot of offline data to initialize the policy, and distribution shift can easily lead to performance degradation during online fine-tuning. To solve the above problems, we propose a learning from demonstrations method named A-SILfD, which treats expert demonstrations as the agent's successful experiences and uses experiences to constrain policy improvement. Furthermore, we prevent performance degradation due to large estimation errors in the Q-function by the ensemble Q-functions. Our experiments show that A-SILfD can significantly improve sample efficiency using a small number of different quality expert demonstrations. In four Mujoco continuous control tasks, A-SILfD can significantly outperform baseline methods after 150,000 steps of online training and is not misled by imperfect expert demonstrations during training.
\end{abstract}


\newcommand{\BibTeX}{\rm B\kern-.05em{\sc i\kern-.025em b}\kern-.08em\TeX}

\begin{document}


\pagestyle{fancy}
\fancyhead{}


\maketitle 


\section{Introduction}
\label{Section1}
Deep reinforcement learning (DRL) for sequential decision problems has shown significant advantages, with excellent performance in robot control \cite{andrychowicz2020learning, ji2022hierarchical}. However, building a control policy based on DRL requires a fine-grained design of the training process in order to explore the environment. The DRL agents must struggle with high sample complexity for a long time, hindering the development of DRL in robot control. Therefore, researchers have proposed two approaches for building a DRL control policy. On the one hand, some methods train the control policy in a simulation environment with unlimited data sources and perform the sim-to-real transfer \cite{zhao2020sim}. On the other hand, many methods use offline data collected from previous training to guide exploration and improve sample efficiency.

There are several methods to use expert demonstrations. In imitation learning (IL), the agents do not need environmental rewards but learn policy by imitating the teacher's behavior \cite{DBLP:conf/nips/HoE16, DBLP:conf/corl/GhasemipourZG19, DBLP:conf/iclr/KostrikovADLT19, DBLP:conf/iclr/ReddyDL20, DBLP:conf/iclr/BrantleySH20}. Therefore, IL requires high-quality expert demonstrations. In practice, expert demonstrations may be generated by human or simple control methods, and the quality of expert demonstrations varies. IL performs poorly in imperfect expert demonstrations. Learning from demonstrations (LfD) also uses expert demonstrations, combining the advantages of IL and RL to improve the sample efficiency of online training \cite{ramirez2022model}. Existing LfD methods enhance RL by either putting expert demonstrations into a replay buffer for value estimation or using expert demonstrations to pre-train the policy \cite{vecerik2017leveraging, hester2018deep}. Unfortunately, the utilization of expert demonstration is inefficient. Other LfD methods are based on on-policy RL algorithms that add a demonstration-guided term to guide policy improvement \cite{cruz2017pre, kang2018policy, DBLP:conf/iclr/RengarajanVSKS22}. Offline RL uses a large number of expert demonstrations to learn the policy without interacting with the environment \cite{DBLP:conf/nips/KumarFSTL19, DBLP:conf/icml/FujimotoMP19, fujimoto2021minimalist, kumar2020conservative}. Offline to Online RL combines offline training and online fine-tuning, enabling policy to achieve better performance \cite{nair2020awac, DBLP:conf/corl/LeeSLAS21, DBLP:conf/iclr/KostrikovNL22, zhao2021adaptive}. However, both types of methods require a large amount of offline data to train the policy, which cannot be effective under limited expert demonstrations. In addition, Offline to Online RL methods faces severe Q-function estimation errors due to distribution shift.

Noting the limitations of existing methods, we propose a LfD method, \textbf{A}ccelerating \textbf{S}elf-\textbf{I}mitation \textbf{L}earning \textbf{f}rom \textbf{D}emonstrations (A-SILfD). It is based on the Actor-Critic update framework and more effectively uses a small number of expert demonstrations to improve sample efficiency.

The main contributions of this paper are as follows.

\begin{table*}[htbp]
\centering
\small
\caption {Comparison of different methods.}
\label{tab2} 

\begin{tabular}{ccccccccc}
\toprule
Method               & \begin{tabular}[c]{@{}c@{}}Environmental\\ Reward\end{tabular} & \begin{tabular}[c]{@{}c@{}}Trajectory\\ Reward\end{tabular} & \begin{tabular}[c]{@{}c@{}}Number of \\ Trajectories\end{tabular} & \begin{tabular}[c]{@{}c@{}}Offline \\ training\end{tabular} & \begin{tabular}[c]{@{}c@{}}Trajectory \\ Evaluation\end{tabular} & \begin{tabular}[c]{@{}c@{}}Policy \\ Constraints\end{tabular} & \begin{tabular}[c]{@{}c@{}}Sample\\ Efficiency\end{tabular} & Problem                    \\ \midrule
IL                   & \ding{53}                                                      & \ding{53}                                                   & Small                                                             & \ding{53}                                                   & \checkmark                                                       & \ding{53}                                                     & Low                                                         & Misleading                 \\
LfD                  & \checkmark                                                     & \checkmark                                                  & Small                                                             & Not True                                                  & \checkmark                                                       & \checkmark                                                    & Medium                                                      & Misleading and Utilization \\
Offline to Online RL & \checkmark                                                     & \checkmark                                                  & Large                                                             & \checkmark                                                  & \ding{53}                                                        & \checkmark                                                    & High                                                        & Distribution Shift    \\ \bottomrule    
\end{tabular}
        
\end{table*}

\begin{itemize}
\item \textbf{Adaptive adjustment of experience replay buffer}: We add expert demonstrations to the experience replay buffer and replace old data with better experiences during training to avoid misleading policy improvement through imperfect demonstrations.
\item \textbf{Full use of experience data}: We use policy constraints to make the agent imitate agent's successful experience and realize the full use of data in the experience replay buffer.
\item \textbf{Ensemble Q-functions}: By introducing randomness through ensemble learning, our method smoother the policy improvement process and avoids the performance degradation caused by distribution shift.

\end{itemize}

We conducted experiments in four Mujoco continuous control tasks using expert demonstrations of different quality (expert demonstrations, mixed expert demonstrations, and imperfect expert demonstrations). The same demonstrations were used for all methods in the experiments. Our experimental results show that our methods are not affected by the quality of the expert demonstrations. Furthermore, our method outperforms all baseline methods regarding sample efficiency and final performance for different quality of expert demonstrations. Moreover, we performed an ablation experiment to analyze the impact of each component on the algorithm.

\section{Related Work}
\label{Section2}

\subsection{Imitation Learning}
IL aims to train a policy to mimic an expert policy as closely as possible and does not use environmental rewards. The simplest IL method is behavioral cloning (BC) \cite{DBLP:conf/mi/BainS95}, which uses a supervised learning approach to fit expert demonstrations. GAIL \cite{DBLP:conf/nips/HoE16} is a classical adversarial-based IL method that uses discriminators to classify expert and sampled data. Besides, AIRL \cite{fu2017learning}, IC-GAIL \cite{wu2019imitation}, WGAIL \cite{wang2021learning} and F-IRL \cite{DBLP:conf/corl/NiSWGLE20} are also adversarial-based IL methods. DAC \cite{DBLP:conf/iclr/KostrikovADLT19} extended IL to the Actor-Critic methods. OPOLO \cite{zhu2020off} emphasizes imitating only expert states. Self-Adaptive Imitation Learning (SAIL) \cite{zhu2022self} uses sub-optimal expert demonstrations to adaptively adjust the teacher buffer to bring the policy close to and beyond the expert's policy. IL is limited by the expert demonstrations' quality and requires additional sample data to learn the discriminator, resulting in a less efficient sample for the method.

Self-imitation learning (SIL) \cite{oh2018self} is different from imitation learning. SIL is a method that uses the agent's successful experience to encourage the agent to explore and improve the sample efficiency. Our work draws on the idea of SIL.

\subsection{Learning from Demonstrations}
Unlike IL, the idea of LfD is to guide online learning with the help of expert demonstrations, which require environmental rewards. DDPGfD \cite{vecerik2017leveraging} first uses expert demonstrations and training policy, followed by online fine-tuning. DQfD \cite{hester2018deep} trains the Q-network with expert demonstrations to achieve an accurate Temporal-Difference (TD) error at the beginning of the interaction between the agent and the environment. POfD \cite{kang2018policy} uses the idea of GAIL to reshape a reward function by minimizing the occupancy measure of agents and experts. LOGO \cite{DBLP:conf/iclr/RengarajanVSKS22} uses demonstrations to guide exploration during the policy guidance phase, but it is based on the on-policy method——TRPO \cite{schulman2015trust}, resulting in a less efficient sample. LfD can learn effective policy faster by using fewer expert demonstrations in combination with online learning, but the pre-trained policy may also perform poorly in online learning. Our method allows for more efficient use of imperfect expert demonstrations.

\subsection{Offline RL and Offline to Online RL}
Offline RL uses a large amount of pre-collected offline data to train the policy without online interaction. CQL \cite{kumar2020conservative} adds a regularization term to the update of the Q-function so that the Q-function can conservatively estimate a lower bound on the Q-value. TD3-BC \cite{fujimoto2021minimalist} avoids the distribution shift problem by adding additional BC loss in the policy improvement.

Offline to Online RL emphasizes offline training and online fine-tuning. However, a severe bootstrapping error is triggered due to the distribution shift during the online training, which destroys a good policy for Offline RL training. AWAC \cite{nair2020awac} solves the distribution shift by adding regular constraints to the policy. Balanced Replay \cite{DBLP:conf/corl/LeeSLAS21} sets different priorities for online and offline data. IQL \cite{kostrikov2021offline} uses the SARSA style to reconstruct the value function. Offline to Online RL is ineffective when there are few expert demonstrations. Moreover, online training can easily cause the offline policy to collapse. In contrast, our method does not have offline training, directly uses a small number of expert demonstrations to guide online training, and is less affected by distribution shift.

Table~\ref{tab2} compares the differences between the three kinds. The main problems can be summarized as imperfect demonstrations misleading policy improvement, low utilization of demonstrations, and performance degradation due to distribution shift. In this work, we attempt to fill the aforementioned research gaps.
\vspace{-0.2cm}

\section{Leverage the Challenges of Expert Demonstrations}
\label{Section3}

In this section, we have compiled the challenges that can be faced using expert demonstrations.

\subsection{Problem Setting}
\label{Section3.1}

In real-world robotic tasks, a small number of trajectories can be collected by human experts or simple controllers (e.g., PID controllers), which we call expert demonstrations, defined as $\zeta  = \left \{ \varsigma_1,\varsigma_2 ,\dots,\varsigma_n   \right \}$, where $\varsigma_i = \left ( s_t,a_t,r_t,s_{t+1} \right )_{t=1}^{T}$ and $i = 1 \sim n$. The quality of these trajectories can be judged using the reward function. Due to the difficulty of collecting data in the real world, our goal is to fully use a small number of trajectories $\varsigma$, combined with online training to improve the sample efficiency. Eventually, the agent outperforms the imperfect sampling policy and approaches the optimal policy $\pi^{*}$. We mainly consider off-policy RL algorithms, which can train the agent with online and offline data. In particular, the Actor-Critic method can make full use of off-policy data. The two easiest ways to use expert demonstrations are as follows. (1) Use BC to pre-train the policy and fine-tune the policy with online training. (2) Populate the offline data directly into the replay buffer and train the policy. However, these methods have the following disadvantages: (1) They do not take full advantage of the demonstrations; (2) The demonstrations may be imperfect, leading to negative guidance during online training \cite{nair2020awac}.

\subsection{Imperfect Expert Demonstrations}
\label{Section3.2}

\begin{figure}[htbp] 
\centering  
\includegraphics[width=8.5cm]{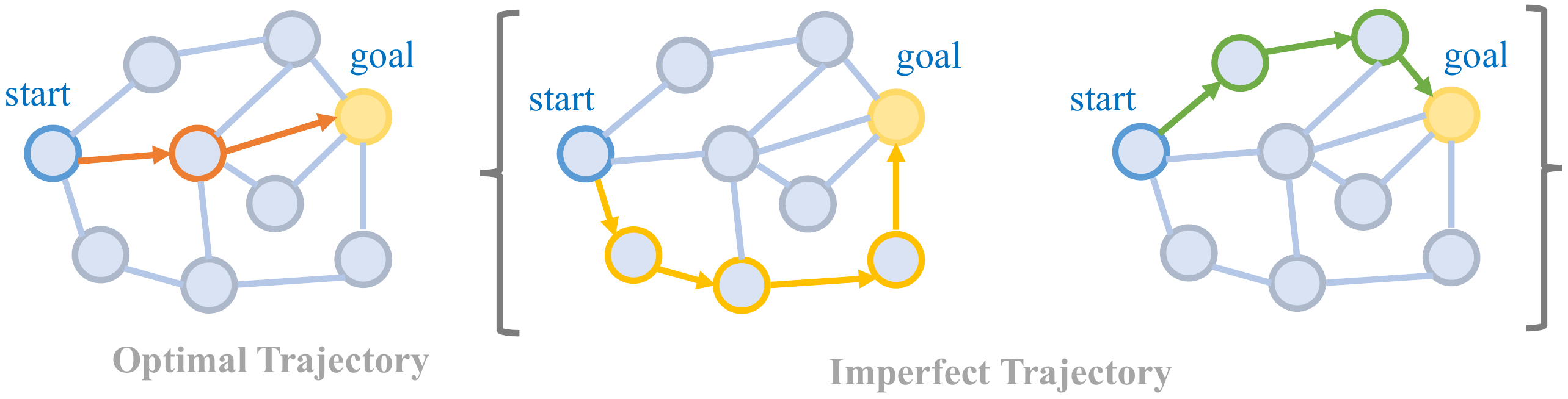}  
\caption{Schematic diagram of optimal and imperfect trajectories.}
\label{fig1}
\end{figure}

As shown in Figure~\ref{fig1}, the left side represents the best trajectory, while two imperfect expert trajectories appear on the right. Suppose the policy is initialized directly using imperfect expert demonstrations. Then, when using this policy for decision-making, the policy directs the agent to reach the $goal$ along the sub-optimal route.

\vspace{-0.3cm} 
\begin{figure}[htbp] 
\centering  
\includegraphics[width=7cm]{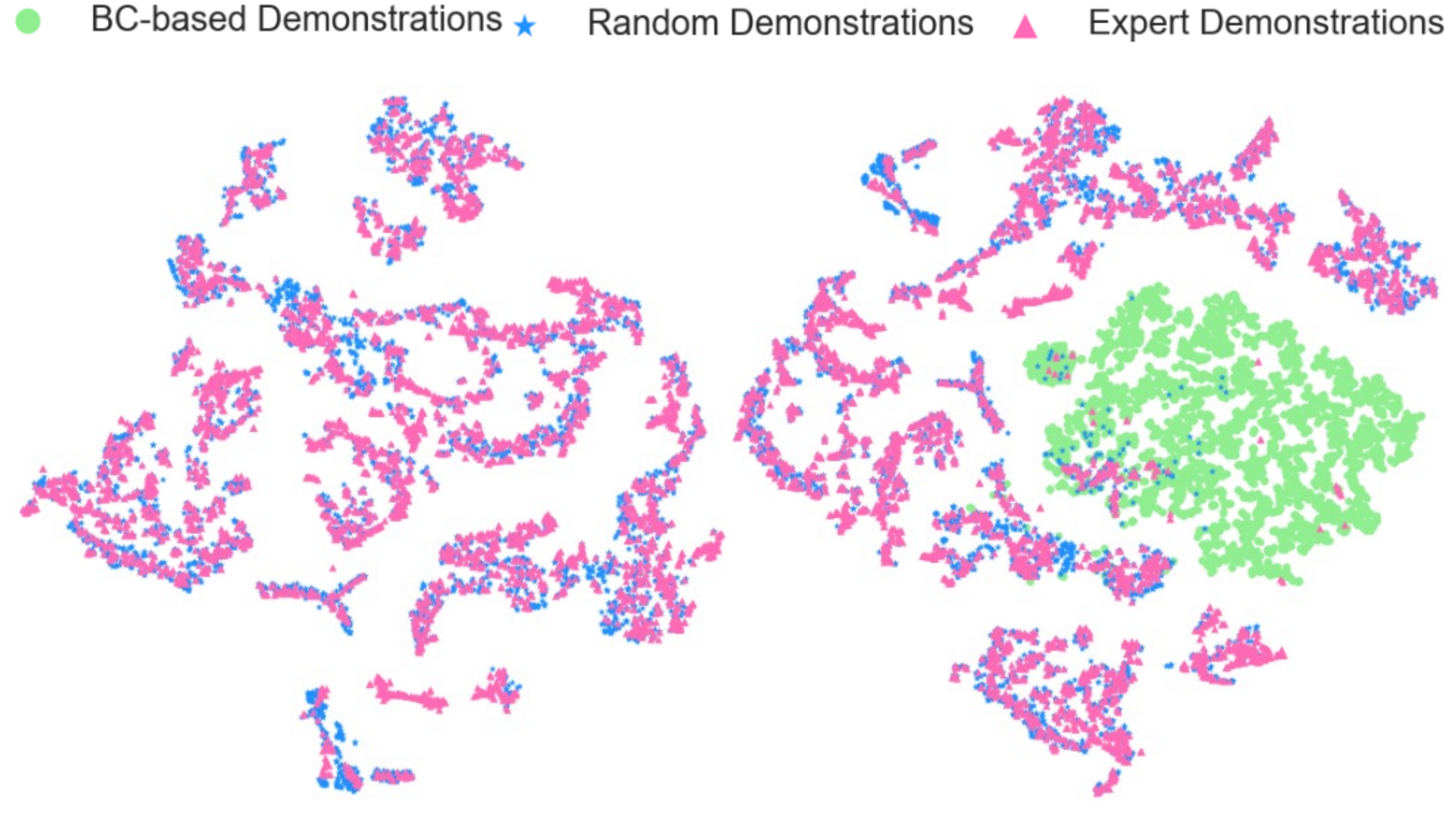}  
\caption{The distribution of different data in the state space. We use t-SNE to downscale the states of Hopper-v2 to the two-dimensional plane.}
\label{fig2}
\end{figure}
\vspace{-0.3cm}

In the continuous state space, the influence of imperfect expert demonstrations remains. The blue area in Figure~\ref{fig2} shows the state distribution of the expert demonstrations, and the pink area shows the state distribution of the data sampled using the BC policy. The demonstrations collected by the pre-trained policy will overlap with the expert demonstrations in the state space. As a result, imperfect demonstrations will be used to update the policy. This negative guidance may move the agent away from the state space where the optimal trajectories are located, thus affecting the agent's exploration.

\subsection{Distribution Shift and Bootstrapping Error Accumulation}
\label{Section3.3}

If we want to explore the unknown region, we need to increase the randomness of the policy. As shown on the right side of Figure~\ref{fig2}, the blue area is the state of the expert demonstrations, while the green area is the state of the sampled data of the policy with more randomness. While random data may start with the same state as the expert demonstrations, actions with greater randomness may result in subsequent states that deviate significantly from the expert demonstrations. Due to the distribution shift, the data covers different state-action regions. Therefore, the Q-function cannot provide accurate value estimates for such out-of-distribution (OOD) samples. Severe bootstrapping errors can destroy the good initial policy obtained by pre-training.

\begin{figure}[htbp] 
\centering  
\includegraphics[width=8cm]{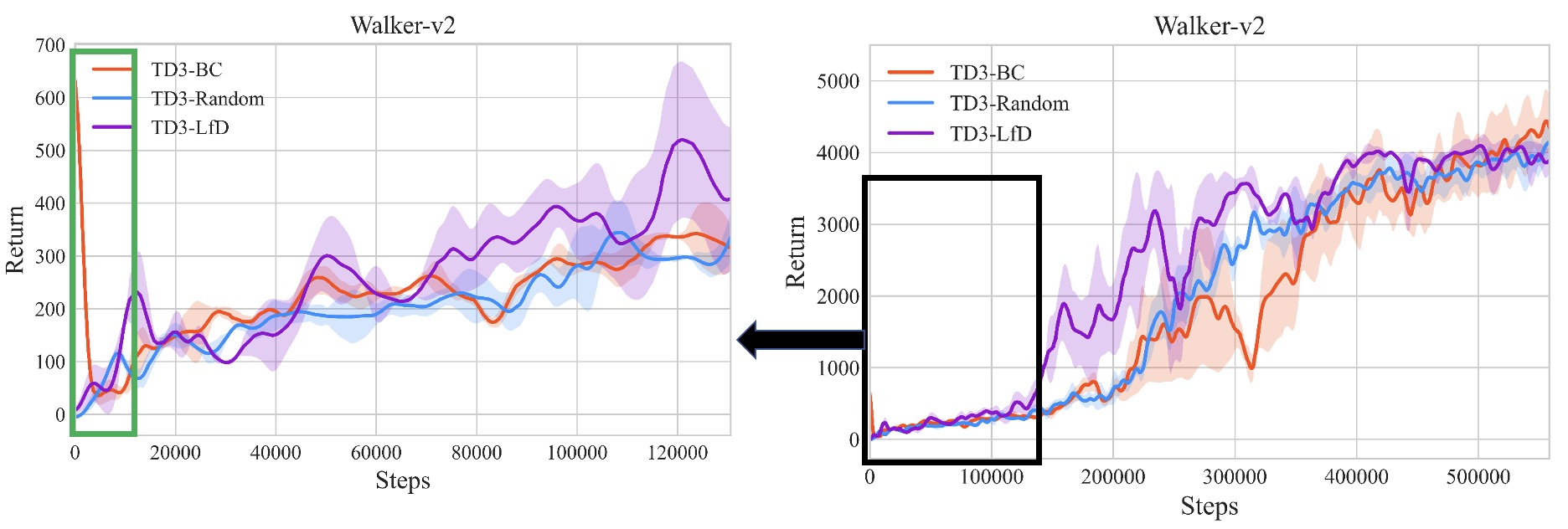}  
\caption{Experiments on Walker2d-v2. TD3-Random denotes the TD3 algorithm trained from scratch. TD3-BC denotes the TD3 algorithm using a BC initialization policy. TD3-LfD denotes the TD3 algorithm based on the LfD idea.}
\label{fig3}
\end{figure}

Let us analyze what the errors associated with the Q-function are. Let the actual Q-value for the current $k$ iterations be $Q^\pi(s,a)$ and the estimated Q-value be $Q_k(s,a)$, and the estimation error can be expressed as $\Gamma_k = \left | Q_k(s,a) - Q^{\pi}(s,a) \right | $.

The current Bellman error is $\delta_k(s,a)\ =\ \left|Q_k(s,a)\ -\ {\hat{Q}}_{k-1}(s,a)\right|$, where ${\hat{Q}}_{k-1}(s,a)\ =\ r(s,a)\ +\gamma \mathbb{E}_{T(s^\prime|s,a)}\left [ \max_{a^{\prime }}{\hat{Q}}_{k-1}(s^\prime,a^\prime) \right ] $.

We can obtain 
\begin{equation}
\Gamma_k(s,a)\ <=\ \delta_k(s,a)\ +\ \gamma\max_{a^\prime}\mathbb{E} _{T(s^\prime|s,a)}\left [ \Gamma_{k-1}(s,a)   \right ] 
\end{equation}

We want $\delta_k(s,a)$ to be as large as possible because $\delta_k(s,a)$ affects the update of the Q-function, which helps to fit the out-of-distribution state-action pairs and reduces the estimation error. However, in practice, out-of-distribution samples lead to a large estimation error $\Gamma_k$, accumulating in each iteration. Thus it can lead to the training policy that deviates significantly from the optimal policy.

Figure~\ref{fig3} shows the problem caused by the distribution shift. The right panel shows the reward curve for the evaluation phase of the training process, and the left panel captures the first 120,000 steps of the reward curve. If the BC initialization policy is used, it leads to a large estimation error $\Gamma_k$ of the Q-function. The estimation error accumulates in the early stage of training, thus distorting the pre-trained policy. As a result, the agent performance may increase degradation rapidly (as in the green box on the left side of Figure~\ref{fig3}). Although the idea of LfD can be used for pre-training, the estimation error $\Gamma_k$ of the Q-function in the unknown region will also be larger, resulting in the pre-training policy not achieving the expected results.

\section{Method}
\label{Section5}

\begin{figure}[htbp] 
\centering  
\includegraphics[width=8cm]{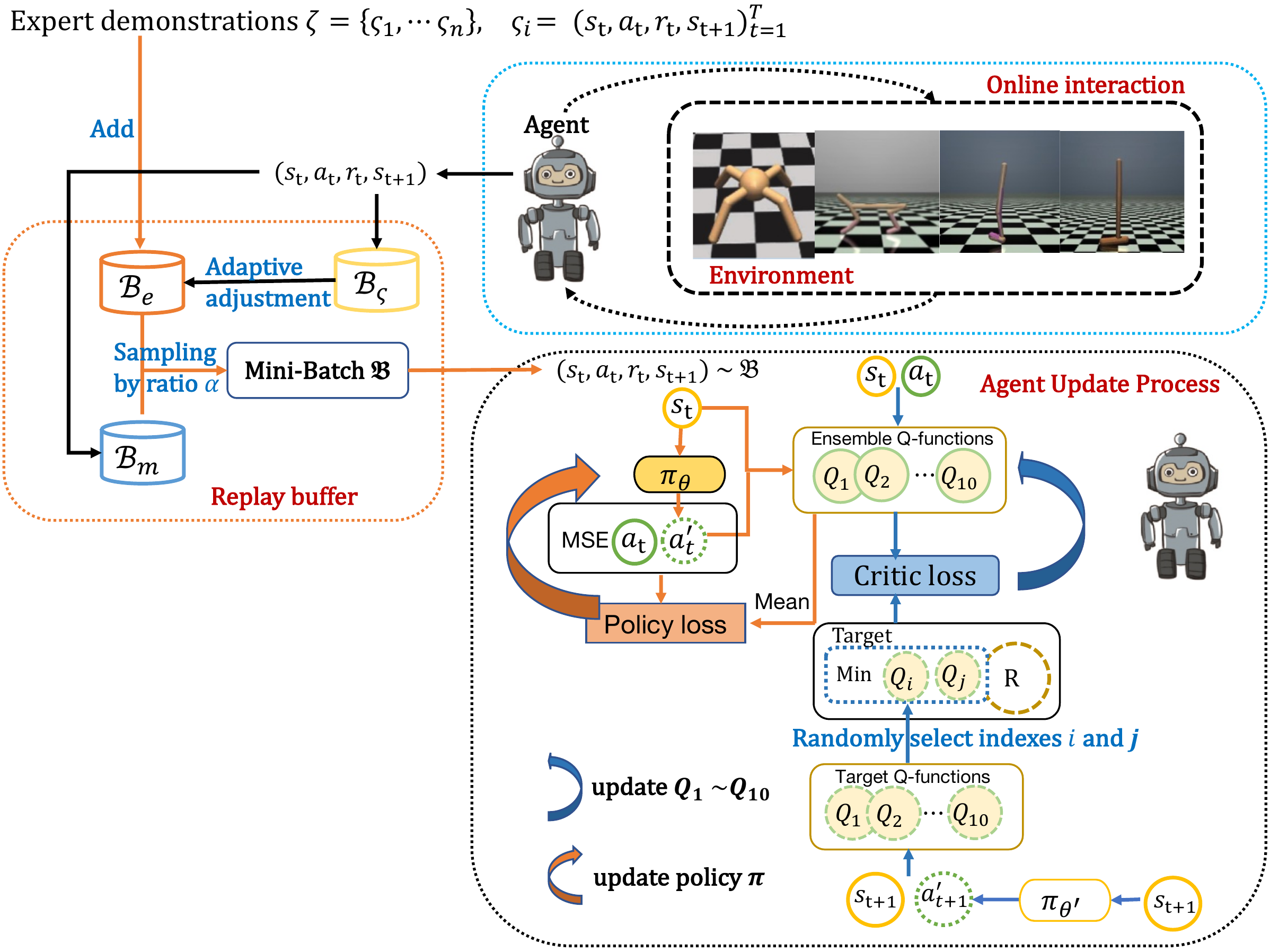}  
\caption{The training flow of A-SILfD.}
\label{flow}
\end{figure}

In this section, we describe A-SILfD. Figure~\ref{flow} illustrates the training process of A-SILfD. First, Section~\ref{Section4.2} describes how the expert demonstrations are used and how the experience replay buffer is adaptively adjusted. Then, in Section~\ref{Section4.3}, we introduce our policy constraints method. Finally, Section~\ref{Section4.4} describes how to train Ensemble Q-functions and policy.

\subsection{Preliminaries}
\label{Section4.1}

RL is typically modeled with Markovian Decision Processes (MDP) and is defined as a five-tuple $M=\left \langle \mathcal{S},\mathcal{A},T,r,\gamma  \right \rangle $, where $\mathcal{S},\mathcal{A}$ represent state and action spaces; $T(s^{\prime}|s,a)$ is the state transfer probability; $r(s,a)$ represents the reward function; $\gamma \in [0,1]$ is the discount factor \cite{andrew1998reinforcement} indicating the influence of future reward at the current time.

The goal of RL is to obtain a policy $\pi$ that maximizes the cumulative (discounted) reward:

$$
\setlength{\abovedisplayskip}{1pt}
\setlength{\belowdisplayskip}{1pt}
\pi^{*} = arg\max_{\pi} \mathbb{E} _{\pi}\left [ \sum_{t=0}^{\infty  }\gamma^{t}r(s_t,a_t)  |a_t = \pi(\cdot |s_t) \right] $$

The Actor-Critic methods learn both the value function and the policy function. Twin Delayed Deep Deterministic policy gradient algorithm (TD3) \cite{fujimoto2018addressing} is an off-policy Actor-Critic method that learns Q-function $Q_{\phi}(s,a)$ parameterized by $\phi$ and a deterministic policy $\pi_{\theta}$ modeled as a feedforward network, parameterized by $\theta$. TD3 alternates between critic and actor updates by minimizing the following objectives respectively: 

\vspace{-0.3cm} 
$$
\setlength{\abovedisplayskip}{1pt}
\setlength{\belowdisplayskip}{1pt}
\mathcal{L}^{critic}(\phi) = \mathop{\mathbb{E}}\limits_{(s,a,s^{\prime } \in \mathcal{B} ) } \left [ Q_{\phi}(s,a)-r-\gamma  Q_{\phi^{\prime }}(s^{\prime } ,\pi_{\theta}(s^{\prime})  \right ] $$
\vspace{-0.1cm} 
$$
\setlength{\abovedisplayskip}{1pt}
\setlength{\belowdisplayskip}{1pt}
\mathcal{L}^{actor}(\theta) =  \mathop{\mathbb{E}}\limits_{s  \in \mathcal{B} ,a \sim \pi_{\theta}  }\left [ -Q_\phi(s,a) \right ]  $$

where $\mathcal{B}$ is the replay buffer and $\phi^{\prime}$ is the target parameters .

\subsection{Self-Imitation Learning from Demonstrations}
\label{Section4.2}
We propose A-SILfD, which can use expert demonstrations and the agent's successful experiences. Our method differs from existing LfD methods in that we consider the expert demonstration as the agent's successful experience as a reference for the agent to learn.

Our method has two replay buffers. First, we store expert demonstrations and successful experiences into the experience replay buffer $\mathcal{B}_{e}$. We store transitions sampled by the agent into the sample replay buffer $\mathcal{B}_m$. 

For the deterministic policy, if the cumulative reward of the trajectory generated by the policy $\pi_i$ is high, it is closer to the optimal policy $\pi^\ast$. As the policy is updated, trajectories with high cumulative reward should be fully used as successful experiences. In addition, we use the cumulative reward of the whole trajectory as the basis for judging the quality of the trajectory, which can effectively avoid the influence of locally optimal actions. Locally optimal actions may lead to a high reward for a particular step, but a high cumulative reward means a high quality of the whole trajectory.

\subsubsection{Adjustment of experience replay buffer $\mathcal{B}_{e}$.}

We record the cumulative reward list $r_{sum} = \{r_{\varsigma_1},r_{\varsigma_2},...,r_{\varsigma_n}\}$ of the trajectories in $\mathcal{B}_{e}$ and derive the minimum cumulative reward $r_{min}(\varsigma )$. During the training process, we store the transitions of each episode in the trajectory replay buffer $\mathcal{B}_{\varsigma}$. At the end of each episode, we calculate the cumulative reward $r_e(\varsigma)$ for transitions in $\mathcal{B}_{\varsigma}$. If $r_e(\varsigma)>r_{min}(\varsigma )$, we add all transitions in $\mathcal{B}_{\varsigma}$ to $\mathcal{B}_{e}$, update the cumulative reward list $r_{sum}$ and recalculate $r_{min}(\varsigma )$. By tuning $\mathcal{B}_{e}$, we ensure that the quality of trajectories in $\mathcal{B}_{e}$ increases gradually as the policy improvement, avoiding misleading policy improvement by imperfect experience.

\vspace{-0.2cm} 
\subsubsection{Usage of $\mathcal{B}_{e}$ and $\mathcal{B}_m$.}

Since $\mathcal{B}_{m}$ stores all transitions sampled by online interactions, and $\mathcal{B}_{e}$ stores the agent's successful experiences, some successful experiences will exist in $\mathcal{B}_m$ and $\mathcal{B}_{e}$. Since the size of $\mathcal{B}_{e}$ is much smaller than that of $\mathcal{B}_m$, when sampling both proportionally, the successful experiences will have a higher probability of being sampled as training data, ensuring the full use of experiences.

\subsection{Policy Constraints}
\label{Section4.3}

The goal of RL is to maximize the expected cumulative (discounted) rewards $\eta (\pi_{\theta})=\mathbb{E} _{\pi_{\theta}}[\sum_{t}^{\infty}{\gamma^t r(s_t,a_t)}]$ along the entire decision process under the current policy $\pi_{\theta}$.

We have an experience replay buffer $\mathcal{B}_e$ and a sample replay buffer $\mathcal{B}_m$. We use $d_{\pi}(s,a)$ to denote the state-action distribution under the policy $\pi$. Moreover, we denote the state-action distribution of the samples in $\mathcal{B}_{e}$ and $\mathcal{B}_{m}$ by $d_{e}(s, a)$ and $d_{m}(s, a)$.

We expect the $d_{\pi}(s,a)$ and $d_{e}(s, a)$ to be close as the policy is learned. In addition, we need to avoid over-exploration of the action. Over-exploration actions can cause the next state to be shifted far from the known region. So, we represent the policy improvement as the following constrained optimization problem.

\begin{equation}
\begin{split}
\label{eq6}
\min_{\theta}{J(\theta)}=-\eta\left(\pi_\theta\right) \quad s.t.\quad  \quad  \mathbb{D} [d_{\pi_{\theta}}(s,a)||d_{e}(s,a)] <  \kappa    
\\  and \quad \quad  \mathbb{D} [d_{\pi_{\theta}}(s,a)||d_{m}(s,a)] <  \kappa     
\end{split}
\end{equation}

$\mathbb{D}$ denotes a measure of distribution distance. $\mathbb{D}\left [ d_\pi(s,a)|| d_{e}(s,a) \right ]  < \kappa$ encourages the state-action distribution of the current policy $\pi$ to match the agent's successful experience state-action distribution. $\mathbb{D}[d_\pi(s,a)||d_m(s,a)] <  \kappa$ seems counterintuitive at first glance, but we aim to avoid exploring too much.

In practice, we avoid explicitly dealing with the inequality constraint by solving the dual problem of Equation \ref{eq6} to solve for Equation \ref{eq6}. Namely, we consider the Lagrangian
\begin{equation}
\label{eq7}
\mathcal{L} (\theta,\lambda)\ =\ -\eta\left(\pi_{\theta}\right)\ +\lambda(\mathcal{G}  (\pi_{\theta}) - 2\kappa )\ 
\end{equation}
where $\mathcal{G}  (\pi_{\theta}) = \mathbb{D}[d_{\pi_{\theta}}(s,a)||d_{e}(s,a)] +  \mathbb{D}[d_{\pi_{\theta}}(s,a)||d_{m}(s,a)] $ and solve the problem $\min\limits_{\theta}\max\limits_{\lambda>0}{\mathcal{L} (\theta,\lambda)}$ by alternately optimizing $\theta$ and $\lambda$.

Optimizing $\lambda$ with fixed $\theta$ is equivalent to
\begin{equation}
\label{eq10}
\min_{\lambda>0}\lambda(2\kappa - \mathcal{G}  (\pi_{\theta}) )\ 
\end{equation}
and optimizing $\theta$ with fixed $\lambda$ boils down to solving
\begin{equation}
\setlength{\abovedisplayskip}{0.5pt}
\setlength{\belowdisplayskip}{0.5pt}
\label{eq11}
\min_{\theta}-\eta\left(\pi_{\theta}\right)\ +\lambda\mathcal{G}  (\pi_{\theta}) \ 
\end{equation}
To solve the above problem, we need to adjust to both $\theta$ and $\lambda$.

In the optimization Equation \ref{eq10} process, the constraints requires $\mathcal{G}(\pi_{\theta}) < 2\kappa$. Therefore, if $\mathcal{G}  (\pi_{\theta}) > 2\kappa$, the $\lambda$ must be increased. When $\mathcal{G}  (\pi_{\theta}) < 2\kappa$, the $\lambda$ can be made as small as possible. All $\lambda > 0$ encourage $\mathcal{G}  (\pi_{\theta})$ to decrease gradually. From another point of view, $\mathcal{G}  (\pi_{\theta}) > 2\kappa$ indicates a constraint violation and requires an increase in the lambda to impose a larger penalty.

In the optimization Equation \ref{eq11} process, as the policy continues to converge, the state-action distributions of $d_{\pi_{\theta}}(s,a)$, $d_{e}(s,a)$ and $d_{m}(s,a)$ keep approaching, $\mathcal{G}  (\pi_{\theta})$ gradually decreases.

To facilitate training, we can fix $\lambda$ as a hyper-parameter and turn this problem into a function related only to $\theta$. The optimization problem becomes
\begin{equation}
\label{eq12}
\theta  \gets  arg\min_{\theta}{\mathcal{L} \left(\theta,\lambda\right)} = arg\min_{\theta}{J\left(\theta\right)\ +\lambda\mathcal{G}  (\pi_{\theta})  }
\end{equation}

Since it is difficult to estimate the state-action distribution of each small batch sample during training and impossible to estimate $\mathcal{G}  (\pi_{\theta})$ accurately, we simplify Equation \ref{eq12}. After considering the implications of $\mathcal{G}  (\pi_{\theta})$, we take the mean square error between the predicted action $a ^{\prime} = \pi_{\theta}(s)$ and the actual action $a$ as an approximate estimate of $\mathcal{G}  (\pi_{\theta})$. Since the data of each mini-batch is sampled from two replay buffers, $\mathcal{B}_{e}$ and $\mathcal{B}_{m}$, this approximate estimation can quickly reach the approximation of $\mathcal{G}  (\pi_{\theta})$. The final optimization objective becomes
\begin{equation}
\label{eq13}
\theta  \gets  arg\min_{\theta}  \left [ J\left(\theta\right) +\lambda \mathbb{E}_{\pi_{\theta}} [(a-\pi_{\theta}(s)^2] \right ] 
\end{equation}

We can expand or contract the constraints region by decreasing or increasing the $\lambda$. In the early stage of training, we want the policy $\pi$ to mimic the agent's successful experience as much as possible, so we use a larger $\lambda$. As training continues, we want the agent to explore the unknown state space gradually, so we gradually decrease the $\lambda$, expanding the constraints region.


\begin{algorithm}[htbp] 
\caption{A-SILfD} 
\label{alg:Framwork} 
\begin{algorithmic}[1] 
\REQUIRE ~~\\ 
Initialize experience replay buffer $\mathcal{B}_{e}$, sample replay buffer $\mathcal{B}_{m}$;\\
Initialize ensemble Q-functions parameters $\phi _1,\dots \phi_N$;\\
To achieve the above goals, we choose the mean $\theta$;\\
Initialize target parameters $\theta ^{\prime}\gets \theta$ and $ \phi^{\prime}_{i} \gets \phi_{i}$($i:=1 \sim  N$);\\
Initialize batch size b, experience ratio $\alpha = 0.25$;

\STATE Add the expert demonstrations to $\mathcal{B}_{e}$, record the cumulative reward $\{r_{\varsigma_1},r_{\varsigma_2},...,r_{\varsigma_n}\}$ of the trajectory in $\mathcal{B}_{m}$ and get the minimum cumulative reward $r_{min}(\varsigma )$; 
\FOR{$i=0$ to $n\_episodes$}
\STATE Initialize trajectory replay buffer $R_{\varsigma}$;
\FOR{$t=0$ to $n\_steps$}
\STATE Act $a_t=\pi_{\theta}\left(s_{t}\right)+ \varepsilon,\ \varepsilon\sim clip\left(\mathcal{N}(0,\sigma),-c,c\right)$;
\STATE Observe next state $s_{t+1}$ and reward $r_t$
\STATE Add ($s_t$, $a_t$, $r_t$, $s_{t+1}$,done) to $R_{B}$;
\STATE Add ($s_t$, $a_t$, $r_t$, $s_{t+1}$,done) to $R_{\varsigma}$;
\IF{$len(\mathcal{B}_m) > b$} 
\STATE $\{s_i,a_i,r_i, \dots \}^{\alpha\times b}_{i=1}\sim \mathcal{B}_{e}$, $\{s_i,a_i,r_i, \dots \}^{(1-\alpha) \times b}_{i=1}\sim \mathcal{B}_{m}$
\STATE Update parameters $\phi _1,\dots \phi_N$ using the Equation \ref{eq15};
\STATE Update parameters $\theta$ using the Equation \ref{eq17};
\STATE $\theta ^{\prime}\gets \tau \times \theta + (1-\tau)\times \theta ^{\prime}  $;
\STATE  $ \phi_{i} ^{\prime}\gets \tau \times \phi_{i} + (1-\tau) \times \phi_{i} ^{\prime}$;
\ENDIF 
\ENDFOR
\STATE Calculate the cumulative reward $r_e(\varsigma)$ in $R_{\varsigma}$;
\IF{$r_e(\varsigma) > r_{min}(\varsigma)$} 
\STATE $\mathcal{B}_e \gets \mathcal{B}_e \cup R_{\varsigma}$
\STATE Update the cumulative reward $\{r_{\varsigma_1},r_{\varsigma_2},...,r_{\varsigma_n}\}$ of the trajectory in $\mathcal{B}_{m}$ and get the minimum value $r_{min}(\varsigma)$; 
\ENDIF 
\ENDFOR
\RETURN policy $\pi$; 
\end{algorithmic}
\end{algorithm}

\subsection{Random Selection of Ensemble Q-Functions}
\label{Section4.4}

It is well known that the Actor-Critic method has been suffering from the over-estimation problem \cite{fujimoto2018addressing, haarnoja2018soft}. At the same time, the distribution shift leads to excessive errors in the estimation of the Q-function, and the combination of the two causes further degradation in the performance of the Q-function. To more effectively mitigate the Q-function estimation error due to distribution shift, we use the idea of the REDQ (Chen et al.) \cite{DBLP:conf/iclr/ChenWZR21} to train the ensemble Q-functions.

Our method is updated in the following way. The estimation of the Q-function follows the Bellman Equation and is updated using the Temporal-Difference method. We still use the target network to avoid the over-estimation problem.

Let the parameters of the ensemble Q-functions be $\{\phi_1,\phi_2,...,\phi_N \}$ and the parameters of the target network be $\{ \phi_1^\prime,\phi_2^\prime,...,\phi_N^\prime \}$. The ensemble Q-functions estimates for $ (s^{\prime},a^{\prime})$ are expressed as
           
\begin{equation}
\setlength{\abovedisplayskip}{1pt}
\setlength{\belowdisplayskip}{1pt}
\label{eq14}
Q_\phi(s^{\prime},a^{\prime}) = \min_{ i \in \mathcal{M}  } Q_{\phi_i^\prime}(s^{\prime},a^{\prime})
\end{equation}
where $\mathcal{M}$ denotes a subset of ensemble Q-functions, with the number of elements in the subset $M < N$.

Then the loss of the i-th Q-function is
\begin{equation}
\label{eq15}
\mathcal{L}_i^{critic} = \sum_{(s,a,r,s^{\prime})\in\mathfrak{B}}{(Q_{\phi_i}(s,a)\ - \ r\ -\ \gamma\ Q_\phi(s^{\prime },a^{\prime }))}^2
\end{equation}
where $\mathfrak{B}$ is the transitions of the currently sampled mini-batch and $a^{\prime } = \pi_{\theta}(s^{\prime }) + \varepsilon,\ \varepsilon \sim clip(\mathcal{N}(0,\sigma ),-c,c )$.

We add constraints to the policy improvement to make more effective use of the agent's successful experience in $\mathcal{B}_{e}$ and avoid exploring farther unknown regions. 

The following describes how the policy is updated. Estimation of the Q-value of the current state-action pair $(s,a)$ by ensemble Q-functions.
\vspace{-0.1cm} 
\begin{equation}
\setlength{\abovedisplayskip}{1pt}
\setlength{\belowdisplayskip}{1pt}
\label{eq16}
Q_{\pi_\theta}\left(s,\hat{a} \right)\ =\ \frac{1}{N}\sum_{i=1}^{N}{Q_{\phi_i}(s,\hat{a}})
\end{equation}
where $\hat{a}  = \pi_{\theta}(s)$.

The loss of the policy can be expressed as
\vspace{-0.01cm} 
\begin{equation}
\label{eq17}
\mathcal{L}^{policy} = \sum_{(s,a,r,s^{\prime })\in\mathfrak{B}}{-Q_{\pi_\theta}\left(s,\hat{a} \right) +\lambda(a - \hat{a})^2}
\end{equation}


An et al. use the minimum of N Q-functions as the estimate of Q-value, which can achieve a pessimistic estimation of Q-value and reduce the estimation error. With the gradual increase of N, the difference between the gradients of Q-functions gradually increases, which is the fundamental reason for the improved effect of Q-Ensemble \cite{an2021uncertainty}. However, more than a pessimistic estimation of values in the online training phase can lead to a lack of exploration. Thus, as shown in equation {eq15}, the TD error is derived by randomly sampling a subset of Q-functions from the ensemble Q-functions and computing the minimum of the subset estimates. The trade-off between pessimistic estimation and exploration is made by introducing randomness.

In the policy improvement phase, ensemble Q-functions estimates' average value is used as the current estimated value. This is because the average value represents the intermediate level of the estimation error of the current state-action pair, which can effectively reduce the estimation error.


\section{Experiment Evaluation}
\label{Section5}

We conducted extensive experiments aimed at answering the following questions.
\begin{figure*}[htp] 
\centering  
\includegraphics[width=17cm]{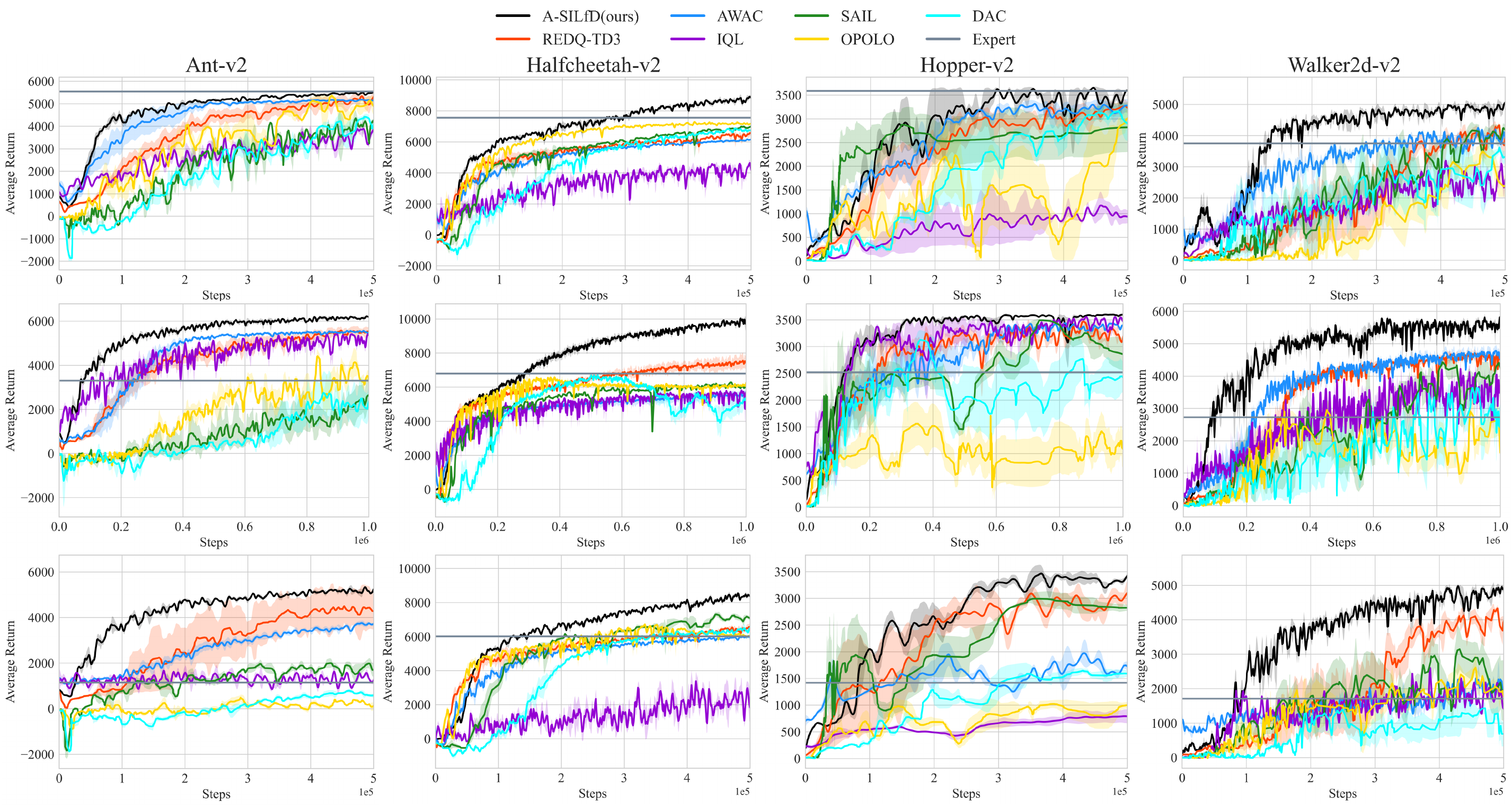}  
\caption{Results for the Mujoco tasks. We conducted experiments on four Mujoco tasks using expert demonstrations (first row), mixed expert and imperfect expert demonstrations (second row), and imperfect expert demonstrations (third row), respectively. The solid lines and shaded regions represent the mean and standard deviation across three runs.}
\label{fig4}
\end{figure*}

\begin{itemize}
\item \textbf{Sample Efficiency}: Does A-SILfD have a higher sample efficiency than other methods (see Figure \ref{fig4})?
\item \textbf{Imperfect Expert Demonstrations}: Can A-SILfD perform better despite imperfect expert demonstrations (see Figure \ref{fig4})?
\item Is A-SILfD more advantageous than other Ensemble Q-functions based methods (see Table \ref{tab1})?
\item What is the impact of each part of A-SILfD on the overall performance (see Figure \ref{fig5} and Figure \ref{fig6})?
\end{itemize}
\subsection{Setup}
\label{Section5.1}

In our experiments, we chose four classic Mujoco control tasks: Ant-v2, HalfCheetah-v2, Hopper-v2 and Walker2d-v2, which can be used as representatives of robot control tasks.

For each task, our expert demonstrations come from the implementation provided by the OPOLO\footnote{https://github.com/illidanlab/opolo-code}, and the imperfect expert demonstrations come from the implementation provided by the SAIL\footnote{https://github.com/illidanlab/SAIL}. All tasks were performed on three random seeds (10, 20, 30) using four trajectories with no more than 1000 steps.

Our method uses a single hidden layer feedforward network as the actor-network and critic-network (Q-function) on all tasks, with 256 neurons. Based on the ensemble learning, we have ten critic networks with the same structure and random initialization, respectively.

All experimental results are evaluated in the corresponding intensive reward environment provided by OpenAI Gym for the final performance. The horizontal coordinates of all figures are the number of steps that have interacted with the environment during training, and the vertical coordinates are the cumulative reward of the current evaluation. The figures allow us to see the sample efficiency of the different algorithms.

\subsection{Sample Efficiency}
\label{Section5.2}

Our baseline methods are shown in Table~\ref{tab-baseline}. All methods are off-policy, based on the TD3 \cite{fujimoto2018addressing} or the Soft Actor-Critic (SAC) algorithm \cite{haarnoja2018soft}, with high data utilization. In particular, REDQ does not require expert demonstrations and is trained from scratch, achieving state-of-the-art results compared to similar methods. 

Due to the limited availability of the LfD methods code, we cannot access all source codes. Moreover, many LfD methods are based on the idea of on-policy, which has low sample efficiency. So we implement REDQ-LfD based on the idea of LfD in combination with REDQ methods and compare them in Section~\ref{Section5.3}.

\begin{table}[htbp]
\centering
\normalsize
\caption {Comparison of baseline methods.}
\label{tab-baseline}
\begin{tabular}{cccc}
\toprule
Algorithm & Types                & Basic algorithm & Year \\\midrule
REDQ \cite{DBLP:conf/iclr/ChenWZR21}      & Scratch              & TD3             & 2021 \\
AWAC \cite{nair2020awac}      & Offline to Online RL & SAC             & 2020 \\
IQL  \cite{DBLP:conf/iclr/KostrikovNL22}     & Offline to Online RL & SAC             & 2022 \\
SAIL \cite{zhu2022self}   & Imitation learning   & TD3             & 2022 \\
OPOLO  \cite{zhu2020off}   & Imitation learning   & TD3             & 2020 \\
DAC \cite{DBLP:conf/iclr/KostrikovADLT19}   & Imitation learning   & TD3             & 2019 \\
\bottomrule
\end{tabular}
\end{table}

REDQ-TD3 is only compared to A-SILfD alone because it does not use expert demonstrations, reflecting the higher sample efficiency that A-SILfD can achieve with a small number of expert demonstrations. Figure \ref{fig4} shows the performance of all methods on Mujoco's four tasks, in which we experimented with expert demonstrations, mixed expert demonstrations, and imperfect expert demonstrations, respectively. In all experiments, our method was not affected by the quality of the expert demonstrations and had a higher sample efficiency.

\subsubsection{Expert demonstrations (Figure \ref{fig4}, first row).}
\label{Section5.2.1}
This part of the experiment uses four expert trajectories. Our method outperforms the baseline methods in all four tasks. AWAC and IQL have a higher initial cumulative reward due to the initial pre-training phase. Affected by the distribution shift, AWAC suffered from a sudden performance degradation at the beginning. The strict policy constraints and the small amount of offline data lead to poor results of the IQL. At the same time, A-SILfD can avoid performance degradation at the initial moment due to the distribution shift. SAIL applies to the case of imperfect expert demonstrations, which perform poorly in this case. Although DAC and OPOLO do not use environmental rewards, the sample efficiency is lower, and the training process is volatile.

\subsubsection{Mixed expert demonstrations (Figure \ref{fig4}, second row).}
\label{Section5.2.2}
This part of the experiment uses two expert trajectories and two imperfect expert trajectories. A-SILfD is ahead of the baseline method on all tasks except the Hopper-v2 task. The IL methods OPOLO and DAC can be misled by imperfect expert demonstrations, leading to poor performance. SAIL does not distinguish well between different quality demonstrations. AWAC and IQL perform better than Section~\ref{Section5.2.1} because demonstrations of varying quality can occupy a larger region of the state space, resulting in better initial results for the Q-function.

\subsubsection{Imperfect Expert demonstrations (Figure \ref{fig4}, third row).}

This part of the experiment uses four imperfect expert trajectories. A-SILfD effectively uses imperfect demonstrations. The IL methods OPOLO and DAC fail to learn effective control policy on Hopper-v2, Ant-V2 and Walker-V2 due to the limitation of the quality of the expert demonstrations. While SAIL can outperform imperfect expert demonstrations, it cannot achieve better performance. AWAC and IQL are only close to the performance of A-SILfD on Hopper-v2 and underperform on other tasks. IQL has no significant performance improvement in the online training. This experiment shows that A-SILfD can better use imperfect expert demonstrations to improve sample efficiency and avoid imperfect data misleading policy improvement.

\subsection{Comparison of Methods based on Ensembles Q-functions}
\label{Section5.3}

In this section, we discuss the effect of ensemble Q-functions. Based on the idea of ensemble Q-functions, we implemented the following method to enhance sample efficiency using expert demonstrations.
\begin{itemize}
\item \textbf{REDQ-LfD}: Due to the limited availability of the LfD methods code, we implemented REDQ-LfD based on the LfD idea. We use the offline data for pre-training before the REDQ - TD3 training and still use the offline data for the subsequent online training. 

\item \textbf{REDQ-BC}: We use BC combined with expert demonstrations to initialize the policy and then train the policy online.
\end{itemize}

\begin{table}[htbp]
\centering
\normalsize
\begin{center}
\caption {Comparison of the method based on the ensemble Q-functions.}
\label{tab1}
\begin{minipage}{0.7\textwidth}
\setlength{\tabcolsep}{1.3mm}{
\begin{tabular}{c|cccc}

\toprule
Benchmark      & Ant          & HalfCheetah       & Hopper       & Walker2d       \\
\midrule
($\mathcal{S},\mathcal{A}$)          & (111,8)          & (17,6)               & (11,3)          & (17,6)            \\\midrule
Basic Reward  & 4600             & 6000                 & 3500            & 3400              \\\midrule
              & \multicolumn{4}{c}{Evaluated Performance/Surpass the Basic Reward} \\\midrule
REDQ      & 1.17/340k        & 1.13/320k            & 0.95/No         & 1.09/410k         \\
REDQ-BC  & 1.20/380k        & 1.25/180k            & 1.01/420k         & 1.20/260k         \\
REDQ-LfD & 1.22/210k        & 1.46/120k            & 1.02/240k         & 1.37/200k         \\\bottomrule
Ours           & \textbf{1.24/180k}        & \textbf{1.48/120k}            & \textbf{1.05/180k}         & \textbf{1.43/130k}        \\ \bottomrule
\end{tabular}}
\end{minipage}
\end{center}
\end{table}

Table~\ref{tab1} shows the results of our experiments. Evaluated Performance indicates the ratio of the cumulative reward to the basic reward at the time of evaluation at 500,000 online interaction steps. Surpass the basic reward indicates the number of interactions required to exceed the basic reward smoothly. On the four Mujoco tasks, the ideas of BC and LfD combined with Ensemble Q-functions can improve the sample efficiency. However, A-SILfD requires fewer steps and has higher sample efficiency than the other methods. In addition, the A-SILfD algorithm still achieves better performance after 500,000 step iterations. The results show that in addition to the idea of ensemble Q-functions, other parts of A-SILfD can effectively improve the algorithm's performance.

\begin{figure}[htbp] 
\centering  
\includegraphics[width=8cm]{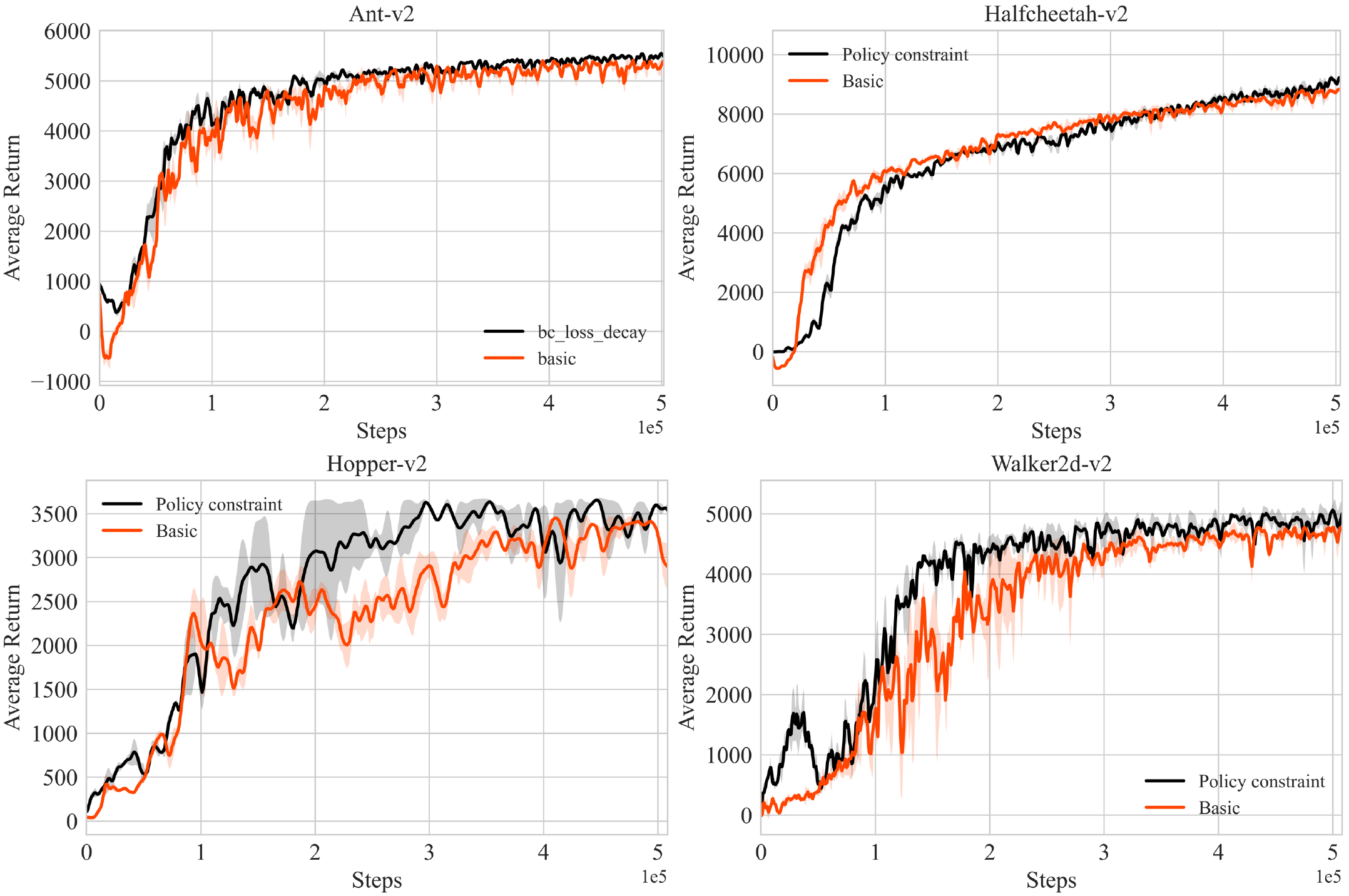}  
\caption{The importance of policy constraints. Basic means no policy constraints is added.}
\label{fig5}
\end{figure}
\begin{figure}[htbp] 
\centering  
\includegraphics[width=8cm]{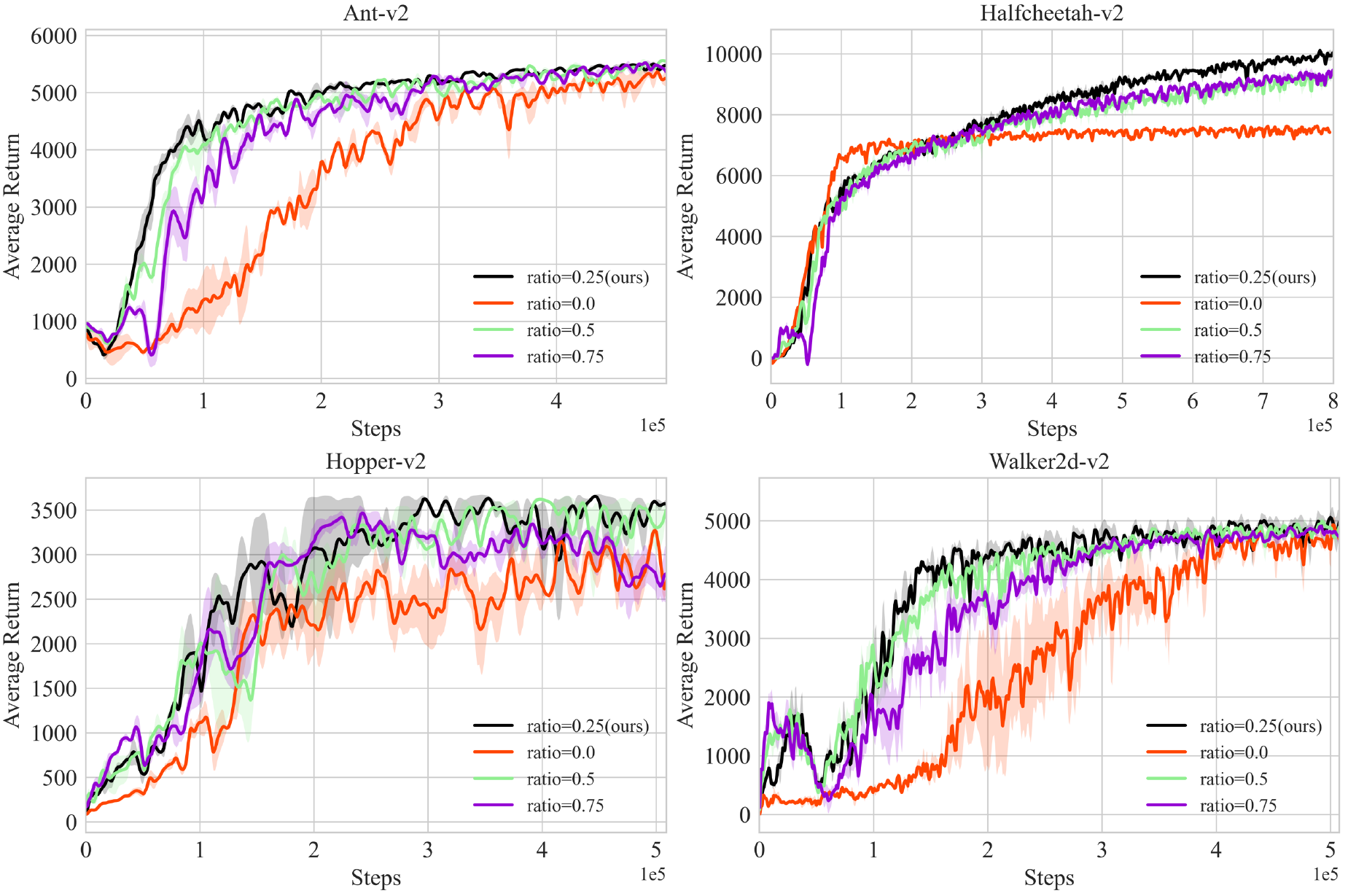}  
\caption{Comparison of different $ratio$ on the four Mujoco tasks.}
\label{fig6}
\end{figure}

\subsection{Ablation Experiment}
\label{Section5.4}

\subsubsection{The Importance of Policy Constraints.}

As shown in Figure~\ref{fig5}, the red curve indicates the A-SILfD's performance without policy constraints, and the black curve indicates the A-SILfD's performance after adding policy constraints. The training data of each mini-batch contains the agent's successful experience and other sampled data. Therefore, the results show that the policy constraint term has the following two effects. (1) Make the policy try to mimic the agent's successful experience so that the algorithm performance reaches the expert's performance more quickly. (2) Constrain the exploration region of the policy to make the algorithm representation more stable. The experimental results show that the policy constraints can effectively improve the performance of A-SILfD.

\subsubsection{Effect of Sampling Ratio.}

When sampling the training data, the data in the sample replay buffer and the experience replay buffer are sampled proportionally. $ratio$ indicates the percentage of data from the experience replay buffer for each mini-batch. We consider four settings. $ratio \in \{0.25,0.5,0.75\}$ means that each mini-batch samples a different proportion of experience data, while $ratio=0.0$ means that the data in the experience replay buffer is not used. As shown in Figure~\ref{fig6}, we can see that $ratio=0.0$ is ineffective, showing the importance of expert demsonstration and the agent's successful experience. The $ratio=0.75$ means that each mini-batch uses more of the agent's successful experience, and the training data lacks exploration, resulting in poor performance. $ratio=0.25$ considers both the imitation of the agent's successful experience and environmental exploration, and we finally choose $ratio=0.25$.

\section{Conclusion}
\label{Section6}
In this paper, we discuss how to effectively use fewer expert demonstrations to improve the sample efficiency of DRL. Considering that the expert demonstrations may be small and imperfect, the existing methods cannot use them effectively. Therefore, we propose an Actor-Critic framework-based LfD method named A-SILfD. A-SILfD stores expert demonstrations as the agent's successful experience in the experience replay buffer. During the training process, A-SILfD evaluates the quality of the trajectory and dynamically adjusts the data in the experience replay buffer. Our experimental results in the Mujoco control task show that. 
\begin{itemize}
\item Our method can improve sample efficiency using different quality expert demonstrations.
\item Through effective self-imitation learning, we can fully use the agent's successful experience to learn the policy so that the policy can eventually achieve higher performance.
\item By ensemble Q-functions, our method makes the policy training process smoother and avoids the performance degradation caused by distribution shift.
\end{itemize}

Although our method could significantly speed up the training process using fewer expert demonstrations, it encountered some limitations. First, there is a challenge in designing the weights for the policy constraints; we set the initial weight to 1 and decay it with the training process. However, the weights must be adjusted separately for different environments to achieve higher performance. Second, although the experimental results surface that the ensemble Q-functions can reduce the estimation error due to distribution shift, we do not provide detailed proof of this point. Finally, since our method relies on the reward function to evaluate the trajectory quality and uses the TD error learning Q-function, it may not work well in a sparse reward setting.




\clearpage


\bibliographystyle{ACM-Reference-Format} 
\bibliography{sample}


\end{document}